\documentclass[twoside,11pt]{article}

\usepackage{jmlr2e}

\usepackage{url}

\hypersetup{
  colorlinks=true,
	citecolor=blue!50!blue,
  linkcolor=blue!50!blue,
  urlcolor=blue!70!blue
}

\usepackage{times}
\usepackage{graphicx} 
\usepackage{rotating}
\usepackage{multirow}
\usepackage{amssymb}
\usepackage{amsmath}
\usepackage{mathtools}
\usepackage{algorithm}
\usepackage{algorithmic}
\usepackage{multirow}
\usepackage{tcolorbox}
\usepackage{tabularx}
\usepackage{array}
\usepackage{colortbl}
\usepackage{fancybox}
\usepackage{tikz}

\tcbuselibrary{skins}

\usepackage[utf8]{inputenc} 
\usepackage[T1]{fontenc}    
\usepackage{hyperref}       
\usepackage{url}            
\usepackage{booktabs}       
\usepackage{amsfonts}       
\usepackage{nicefrac}       
\usepackage{microtype}      
\usepackage{amssymb}
\usepackage{amsmath}
\usepackage{mathtools}
\usepackage{bigdelim} 
\usepackage{algorithm}
\usepackage{algorithmic}
\usepackage{color}
\usepackage{verbatim}
\usepackage{siunitx}

\tcbuselibrary{skins}
\usepackage{fancybox}
\usepackage{tikz}
\usepackage{multirow}
\usepackage{tcolorbox}
\usepackage{tabularx}
\usepackage{array}
\usepackage{colortbl}
\usepackage{caption}
\usepackage{subfig}
\usepackage{wrapfig}
\usepackage{makecell}
\usepackage{float}

\usepackage[symbol]{footmisc}

\usepackage{hyperref}

\tcbset{tab2/.style={fonttitle=\bfseries\footnotesize,fontupper=\footnotesize,
colback=blue!2!white,colframe=blue!75!black,colbacktitle=red!40!white,
coltitle=black,center title,freelance}}

\usepackage{tikz}
\usetikzlibrary{positioning}

\usepackage[font=small,skip=5.5pt]{caption}

\firstpageno{1}

\title{
Synthetic Data: Opening the data floodgates to enable faster, more directed development of machine learning methods}

\author{\name James Jordon \email james.jordon@wolfson.ox.ac.uk \\
	    \addr Department of Engineering Science\\ 
	    University of Oxford,
	    Oxford, United Kingdom  
	    \AND
	    \name Alan Wilson \email awilson@turing.ac.uk \\
	    \addr The Alan Turing Institute, London, United Kingdom
	    \AND
	    \name Mihaela van der Schaar \email mv472@cam.ac.uk \\
	    \addr University of Cambridge, Cambridge, UK\\
	    University of California, Los Angeles, USA \\
	    The Alan Turing Institute, London, UK}

\begin{document}

\maketitle

\begin{abstract}
    Many ground-breaking advancements in machine learning can be attributed to the availability of a large volume of rich data. Unfortunately, many large-scale datasets are highly sensitive, such as healthcare data, and are not widely available to the machine learning community. Generating synthetic data with privacy guarantees provides one such solution, allowing meaningful research to be carried out ``at scale'' - by allowing the entirety of the machine learning community to potentially accelerate progress within a given field. In this article, we provide a high-level view of synthetic data: what it means, how we might evaluate it and how we might use it.
\end{abstract}

\section{Introduction}
In recent times, many ground-breaking advancements have been made in machine learning, such as the well-known use of deep reinforcement learning to play Atari games \cite{atari} and the widespread success of deep learning in general \cite{deeplearnrev}, both of which can be largely attributed to the sheer volume of rich data that is available.

More and more large datasets are becoming available in a wide variety of communities. In the U.S. medical community, for example, the fraction of providers using electronic health records (EHR) increased from 9.4\% in 2008 to 83.8\% in 2015 \cite{ehradoption}. Within the criminal justice system and military, machine learning has the potential to revolutionise crime prevention and detection but the sensitivity of the data involved means that only a small fraction of machine learning researchers will work on developing meaningful and directed methods for these problems. The availability of large datasets presents enormous opportunities for collaboration between data-holders and the machine learning community. Currently, however, the (in)ability to share many of these large datasets is creating a bottleneck on the rate at which cutting edge machine learning methods can be developed and deployed in the real world.

The most common way to mitigate the risk of sharing sensitive data is to de-identify the data - but it is by now well-known that records that have been de-identified can be easily re-identified by linking them to other identifiable datasets \cite{deidentify1,deidentify2,imdb,deidentify5,deidentify6}. As a result of the lack of robustness of existing de-identification methods, new data-sharing policies and regulations have been introduced (such as with GDPR in Europe). Increased control over sensitive data results in greater difficulties for the machine learning research community in accessing said data. As a result, cutting edge methodologies and techniques often take longer to develop and are often not as specialised as they could be for a specific dataset and/or task.

In light of the flaws with de-identification, we might turn to a popular strategy for enforcing privacy known as $k$-anonymisation, in which a dataset is created by "binning" features in such a way that the features for any given sample are also shared by at least $k-1$ other samples. This ensures that any individual is "hidden" among $k$ individuals and thus cannot be identified. Unfortunately, this approach can have serious drawbacks from a utility perspective - if a dataset contains fewer than $k$ outliers in a particular region of the feature space then these outliers will necessarily need to be binned with samples from other parts of the feature space, thus losing a significant amount of information about each sample due to the "wide" bins necessary to achieve this. Even when there are no outliers, the binning process still results in a dataset that contains less information than the original and is typically agnostic to any downstream tasks that might be performed - the information that is thrown away may prove crucial to a given task.

In order for machine learning to realise its enormous potential in solving real world problems, it is key that machine learning researchers are both posed with real questions that need answering, but also data for which those questions can be asked and meaningfully answered. When real data is unavailable or prohibitively slow to gain access to, "synthetic" data offers a promising alternative that can allow researchers to perform meaningful research while waiting for access. In fact, synthetic data can be used by data holders as a means of screening potential machine learning collaborators, selecting only the collaborators who perform best on the synthetic data to give access to the real data.

Unfortunately, generating {\em meaningful} synthetic data is not an easy task and there are several challenges in developing methods for doing so; even defining what it means to be synthetic is non-trivial. In the realm of data release, additional care needs to be taken when attempting to deploy methods that may not perform well - once the (synthetic) data has been released, there is no taking it back, and a lack of understanding of the synthetic data can lead to serious privacy concerns.

\section{What is synthetic data?}
Defining the term synthetic data is itself a difficult task. In the machine learning community, synthetic data often refers to data generated from a set of easy-to-specify distributions that is used to validate a machine learning model. It is most prevalently used in situations where real data does not suffice to validate a model because ground truth knowledge in real data is not attainable (such as in causal inference \cite{ganite} and feature selection \cite{invase} \cite{knockoffgan} \cite{asac} \cite{l2x}). Such datasets are created with the sole purpose of comparing machine learning methods on {\em reproducible} datasets and are {\em not} created to imitate a real dataset. While this type of synthetic data allows for easily reproducible results, the results on the synthetic data do not necessarily reflect the performance of the methods in any real dataset, and so the conclusions that can be drawn from such experiments are limited.

For the sake of data sharing, synthetic data should be generated {\em using a real dataset} to shape the synthetic data. However, now that the "synthetic" data is allowed to depend on the real data, it becomes less clear exactly what is meant by the term synthetic, and how exactly the synthetic data can use the real data while still remaining synthetic. It is crucial that those looking to create synthetic versions of their data understand how a given synthetic data generating model defines synthetic - a naive belief that the term synthetic is equivalent to private can result in serious privacy violations. The popular generative modelling frameworks Generative Adversarial Networks (GANs \cite{original_gan}) and Variational Auto-encoders (VAEs \cite{vae}) use real data during training to train a model capable of generating (or "synthesising") "fake" data points. But what makes these data points "fake"? A quick look at the seminal GAN paper \cite{original_gan} might make it look like a GAN is all we need to generate a synthetic version of a real dataset. However, closer inspection would reveal that nothing is actually being done to prevent the GAN from simply learning to "regurgitate" the real data. In fact, GANs are notorious for suffering from a phenomena referred to as {\em mode collapse} \cite{mode_collapse}, which causes GANs to repeatedly generate data points that are very similar to the most popular training points.

It is for precisely this reason that one needs to be careful when using the phrases "fake data" or "synthetic data". Once we allow for the data generating mechanism to depend on real data, the term synthetic on its own is not enough; it must be accompanied by some sort of constraint that makes clear the limits for how the synthetic data can use the real data. These constraints should be formulated by asking {\em what is it about the real data that makes it unsuitable for release?} This will typically be answered in terms of certain privacy concerns that releasing the real data would have. Formulating these privacy concerns in a rigorous manner is important to allow for methods to {\em provably} satisfy them.

\section{Privacy} \label{sec:priv}
Privacy is an ongoing area of research in the machine learning and statistics communities and has garnered much attention in recent times with the rapid advancement of technologies that allow for easy data collection on large scales. There is no single definition of privacy that will work in all situations. It should also be noted that attaining certain types of privacy on certain datasets may render the resulting synthetic data useless and so either the notion of privacy will need to be revised or the goal of generating synthetic data abandoned. A simple example of this would be the notion that private synthetic data should not allow us to learn any more about an individual than if we had not had the data. Any dataset that allows us to build a good predictive model for a particular feature based on the other features will violate this - if we know the remaining features for a particular person, we can use the predictive model to predict the final feature. More importantly, this can be done {\em even if the person was not in the original dataset}. There are two failures: (1) the privacy notion is not immune to {\em post-processing} and (2) the notion relies on knowing what information is already known about every individual. While the data itself may not reveal anything about a particular individual (it may not even contain the individual), it can be processed (e.g. by being used to train a predictive model) in such a way that by using existing information (such as an auxiliary dataset), more can be learned about an individual.

It is possible to mitigate the risk of (1) by limiting how exactly the synthetic data can be used or accessed (for example by allowing access to it only through a safe environment in which only certain functions can be performed), however, in doing so, we revert back to the problems we had with real data, in that gaining access can be a cumbersome process. As such, we believe that synthetic data is most useful when it allows for {\em public} sharing, without constraint on how the data will be used. It is therefore imperative that we understand not only what it means for the data to be private but also what it means for the data to {\em remain} private after being processed by an adversarial attacker. It is, however, incredibly difficult to demonstrate adversarial robustness empirically; any attempt to do so is naturally limited by the researcher's own capabilities. Instead, the goal should be to obtain {\em provable} guarantees, which make clear assumptions (or preferably no assumptions) about an adversaries capabilities.

Unfortunately, (2) is equally difficult to mitigate. Modelling the knowledge of an adversary is difficult and also risky, since failing to correctly model the full extent of an adversaries possible knowledge can naturally lead to privacy concerns. It is much safer to assume the adversary knows as much as the worst case scenario would be and work from there, though such an approach calls for the strongest privacy guarantees and thus, potentially, the lowest utility from the synthetic data. Moreover, as more and more data (even synthetic data) is released, an adversaries knowledge can easily change. As there is no way to take back data that we release publicly, it is better to be conservative in this regard (such as assuming the adversary is worst-case).

It is our belief that a good, {\em robust} notion of privacy should make very few assumptions about the limitations of a potential adversary. One such notion of privacy is {\em Differential Privacy} \cite{dpbook}. Differential Privacy is a notion of privacy that is immune to post-processing, that is, if something is differentially private, then no matter what is done to the differentially private object (such as a dataset), it will remain differentially private. It also makes no assumptions about any auxiliary data that may (publicly) exist about a given individual. Perhaps, though, the most attractive thing about differential privacy is that it can be defined rigorously, in fairly simple mathematical terms. This means that, for example with synthetic data, a synthetic dataset can be {\em proven} to be differentially private (or not) and it does not need to be assessed empirically.

Of course, there are drawbacks to differential privacy, for example, it requires specification of a parameter, $\epsilon$, that controls "how" private something should be, and determining a good value for $\epsilon$ is not an easy task (because it is not easy to interpret semantically). In some situations, the assumptions (or rather lack of assumptions) can be much stronger than needed, since it relies on determining the worst-case situation and as such data utility often suffers more than is necessary for the task at hand.

Unfortunately, few alternatives to differential privacy actually exist. It has proven to be incredibly difficult to formulate common semantic notions of privacy in a rigorous mathematical way. This typically implies that such notions are not actually achievable in a real sense (such as the notion described at the beginning of this section) or that enforcing such notions would result in zero utility synthetic data.

One such theoretical alternative is membership privacy, introduced in \cite{membership_priv}. Membership privacy is a more abstract notion of privacy than differential privacy, that relies on defining a set of distributions, $\mathbb{D}$, that captures the adversary's possible states of prior knowledge, and a leakage parameter $\gamma$ that determines by how much an adversary's knowledge is allowed to change based on the output of an algorithm. Differential privacy can be recovered from this definition by defining an appropriate $\mathbb{D}$.

\cite{adsgan} propose an empirical notion of private synthetic data which relies on determining how close a synthetic data point is (in Euclidean space) to any real data points.

Investigating existing definitions of privacy and their real-world implications is a crucial step in understanding them and also in convincing communities at large that such definitions are appropriate for the problems they face. While notions such as differential privacy and membership privacy are both theoretical guarantees that need not be tested empirically, the semantics behind their definitions are somewhat unclear and empirical evaluations against real attackers would benefit our understanding greatly. Our recent synthetic data competition \cite{syncomp} pit cutting edge privacy research against real-world attackers to work towards greater clarity over existing privacy methods.

\section{How can synthetic data be used and evaluated?}
While promising from a privacy perspective, it is important to understand the limitations of when and how synthetic data should be used. The question of how to evaluate synthetic data is also closely tied to how it will be used. In the most general case, one might hope for the synthetic data to be a perfect substitute for the real data in anything for which the real data could be used. For this to be true, the distribution of the synthetic data should match the distribution of the real data. Evaluating this empirically, though, is difficult. A lot of time and attention is being given to developing methods for evaluating the similarity of empirical distributions and it is very much an ongoing area of research \cite{kid}.

The risk with using synthetic data for training as a direct replacement for real data is that at run time, the models will be faced with real data. If the distribution of the synthetic data deviates from that of the real data, then distributional mismatch will occur, resulting in phenomena such as {\em covariate shift} \cite{covshift}, leading to sub-optimal model performance. Depending on the severity of the mismatch (which is difficult to quantify due to the lack of good empirical evaluation tools), this could have serious implications.

For this reason, it is best to create synthetic data with a particular use in mind. It might be that we wish to develop an algorithm for learning a predictive model on a given dataset, or that we wish to perform unsupervised clustering, for example. The quality of the synthetic dataset can then be measured with respect to standard metrics used for these tasks by training a model (or a set of models) on synthetic data and testing it (them) on real data (a paradigm referred to as Train on Synthetic, Test on Real (TSTR \cite{tstr})), the performance on the test set can then be reported using standard metrics (such as AUROC) and even compared to the performance achieved when the same model is trained on real data. By comparing the performance difference between training on synthetic and training on real data we can assess how well the synthetic data captures the characteristics of the real data that are important for carrying out the given task.

Unfortunately, TSTR is sensitive to the models chosen for the task; it may be that logistic regression performs similarly on synthetic and real data but that the performance of random forests drops significantly when the synthetic data is used. Aggregating the TSTR values across different models is not as simple as averaging since this can hide the fact that the synthetic data is not appropriate for the training of certain models.

If the synthetic data is intended to be used as a tool for enabling meaningful machine learning research (which we strongly believe is the best use of synthetic data), then it is important that when a comparison of methods is made on the synthetic data, the answer reflects what the answer would be if the same comparison was made on the real data. This comparison may be of two (or more) very different models, or may be of two slight variations of the same model, but in both cases, if the hope is that synthetic data can allow researchers to develop the best methods for us on the real data, then the development process - which will involve many comparisons of this nature - needs to be meaningful on the synthetic data. To this end, we proposed a new metric in \cite{pategan} which we term the Synthetic Ranking Agreement (SRA).

Like TSTR, SRA requires a task to be specified for the data (though like TSTR we could calculate the SRA independently for several different tasks). We train a selection of models on the synthetic data and then test them {\em on the synthetic data}, imitating what a researcher would (have to) do given only the synthetic data. We also train and test the models in the real data. The metric is then defined to be the agreement between the {\em ranking} of the models. While this metric certainly has its limitations (it is limited by the set of models used to calculate it), we believe that metrics such as this are a step in the right direction for enabling the creation of synthetic datasets that will enable more directed machine learning research.

A final, very promising use for synthetic data is in running competitions. Privacy definitions such as differential privacy lend themselves naturally (by varying the privacy parameter, $\epsilon$) to creating a sequence of synthetic datasets with increasingly less and less privacy. At each stage of the competition, researchers can be given a synthetic dataset to develop a method for the given task. The data holder can then evaluate each of the methods on the real data and select the best methods to "advance" to the next round, at which point a new dataset will be released to those still in the competition, this time with a lower privacy guarantee. This sort of "gated" approach can also be used to alleviate the difficulties of granting access to real data. As trust grows within a data-holder and researcher partnership, data with weaker and weaker privacy guarantees can be given to the researcher.

\section{Existing Methods}
DPGAN \cite{dpgan} proposes a framework for modifying the popular GAN framework to be differentially private, relying on the Post-Processing Theorem \cite{dpbook} to change the problem of learning a differentially private generator to learning a differentially private discriminator. Their work uses a technique introduced by \cite{deeplearn} that provides a differentially private mechanism for training deep networks. The key idea is that noise is added to the gradient of the discriminator during training to create differential privacy guarantees. These ideas are also used similarly in \cite{dpganbio}.

PATEGAN \cite{pategan} is similar in spirit; during training of the discriminator differentially private training data is created using the method proposed in \cite{pate1}. The proposed model modifies the PATE framework \cite{pate1,pate2} for use in a generative model setting (specifically for use with GANs). The key to the GAN framework is that the discriminator is a {\em differentiable} module trained to classify samples as either real or generated. The PATE framework provides a differentially private mechanism for classification by training multiple teacher models on disjoint partitions of the data. To classify a new sample each teacher's output is evaluated on the sample and then all outputs are noisily aggregated. This noisy aggregation, though, results in a classifier that is {\em not} differentiable with respect to the parameters of the generator. In order to overcome this problem the student model from \cite{pate1} is used, that involves taking some {\em public} unlabelled data, labelling it using the standard PATE mechanism and then training the student using the resulting labelled data. Because access to any public data is often an unreasonable assumption in synthetic data generation, we adapt this training paradigm in a way that does not require public data by training the student using only outputs from the (differentially private) generator.

While both DPGAN and PATEGAN are promising ideas from a differential privacy perspective, their utility trade-off is significant if one wants to enforce a meaningful level of differential privacy. Moreover, both methods are only evaluated on static data and while both ideas generalise to time-series data, it is not clear how the performance will be affected.

The Simulacrum is a synthetic dataset recently created to replicate the Cancer Analysis System datasets \cite{simulacrum}. To generate this data, the correlation between each pair of variables was calculated, and for each variable, the two variables most correlated with it were selected. A directed graph is then constructed by taking the nodes to be the variables, and the edges to be the selected pairs of (highly) correlated variables. Directions correspond to going from the node with more edges to the node with fewer edges. From this graph, the data is then generated by learning a series of conditional distributions, starting with the root node (whose distribution is not conditional on anything), and then sequentially learning the distributions as we descend the tree, with each variable's conditional distribution being conditioned on its parent nodes. Some clustering and grouping is applied to the conditional distributions to ensure that all conditional distributions correspond to at least 50 records (thus providing 50-anonymity). While the privacy guarantees that are claimed are indeed provided (50-anonymity), the quality of the data is not in any way guaranteed and the decision to use the two most correlated variables and even the same number of variables for each variable is entirely heuristic.

Other existing works generate synthetic data using summary statistics of the original data \cite{summarystats} or based on specific domain-knowledge \cite{diseasespecific}; however, those methods are limited to low-dimensional feature spaces, specific fields and do not provide any theoretical privacy guarantees. \cite{ml4h_current} generates synthetic patient records using a GAN framework. However, \cite{ml4h_current} focuses only on generating discrete variables, whereas DP-GAN and PATE-GAN are both capable of generating mixed-type (continuous, discrete, and binary) variables. \cite{ml4h_current} also does not provide any theoretical privacy guarantees and instead uses ad-hoc notions of privacy which are only validated empirically.

\section{Technical Challenges}
There are several technical challenges that must be overcome in order to develop good synthetic data generation models, which can be broadly separated into 3 main categories:
\begin{itemize}
	\item Data generation
	\begin{itemize}
		\item Generating data for a variety of data types
		\item GANs have significant limitations which need to be addressed
		\item Generating mixed-type data in a non-parametric way is difficult
	\end{itemize}
	\item Privacy
	\begin{itemize}
		\item Mathematical notions need to align with social understanding
		\item Complex data generation methods will require careful application of privacy mechanisms and in some cases will require new machinery
	\end{itemize}
	\item Evaluation
	\begin{itemize}
		\item Empirically comparing distributions is difficult
		\item Existing techniques for evaluating rely on knowing what the synthetic data might be used for
		\item Defining appropriate metrics on complex data structures (e.g. mixed type data) will be necessary
	\end{itemize}
\end{itemize}

\section{The Road Ahead}
While the methods discussed in the previous sections provide promising first steps towards generating private synthetic data, there are still significant limitations and much room for improvement. The differentially private methods, DPGAN and PATEGAN, both rely on different differential privacy enforcing mechanisms, neither of which were designed specifically for synthetic data generation but are for use more generally. Further research can and should be done to refine or replace these mechanisms to be tailored specifically to the problem of data generation.

Moreover, the underlying generative model used in both cases was a GAN, which, while they have proven very successful in the realm of image generation, may not be most appropriate in all settings. Further research should be done to not only improve (or replace GANs) but to understand their limits and potential privacy concerns when used in "vanilla" form (i.e. without additional mechanisms to ensure privacy).

As discussed in Section \ref{sec:priv}, there are several existing definitions of privacy. Some are theoretically grounded such as Differential Privacy, Membership Privacy and k-anonymity (among others) whereas others are defined empirically such as the definition found in \cite{adsgan}. There is a lot of work to be done purely within the realm of privacy, both in understanding the social implications of existing definitions and in developing new frameworks which may align better with social expectations and legal definitions (such as those laid out in GDPR). In fact, anyone working on synthetic data should work closely and discuss with policy experts to really understand what is needed of synthetic data, which will vary between data holders.

Finally, though provable privacy is attractive, we must strive to make it understandable and interpretable. To do so may require empirical evaluations such as our synthetic data competition \cite{syncomp} to shed light on exactly what different notions of privacy might mean in practice. Ultimately, our duty is to those whose data we are collecting, and it is paramount that they understand {\em why} what we are doing to protect their data actually protects their data, so that they trust us with their data, and thus machine learning may continue its rapid advancement.

\medskip
\small
\bibliography{sdm}



\nocite{*}

\end{document}